\documentclass[journal]{IEEEtran}

\ifCLASSINFOpdf
   \usepackage[pdftex]{graphicx}
  
\else
 
\fi

\usepackage{amsmath}

\usepackage{amssymb}

\usepackage{color}
\usepackage{url}
\usepackage{hyperref}
\usepackage{diagbox}

\usepackage{times}
\usepackage{epsfig}
\usepackage{graphicx}
\usepackage{dsfont}
\usepackage{comment}
\usepackage{multirow}

\begin{document}

\title{Learning Compositional Representation for Few-shot Visual Question Answering}

\author{Dalu~Guo and~Dacheng~Tao,~\IEEEmembership{Fellow,~IEEE}% <-this % stops a space
\thanks{D. Guo and D. Tao are with the UBTECH Sydney AI Centre and the School of Computer Science at The University of Sydney, 6 Cleveland Street, Darlington, NSW 2008, Australia (e-mail: dguo8417@uni.sydney.edu.au; c.xu@sydney.edu.au; dacheng.tao@sydney.edu.au).}
}

\markboth{Journal of \LaTeX\ Class Files,~Vol.~14, No.~8, August~2015}%
{Shell \MakeLowercase{\textit{et al.}}: Bare Demo of IEEEtran.cls for IEEE Journals}

\maketitle

\begin{abstract}
Current methods of Visual Question Answering perform well on the answers with an amount of training data but have limited accuracy on the novel ones with few examples. However, humans can quickly adapt to these new categories with just a few glimpses, as they learn to organize the concepts that have been seen before to figure the novel class, which are hardly explored by the deep learning methods. Therefore, in this paper, we propose to extract the attributes from the answers with enough data, which are later composed to constrain the learning of the few-shot ones. We generate the few-shot dataset of VQA with a variety of answers and their attributes without any human effort. With this dataset, we build our attribute network to disentangle the attributes by learning their features from parts of the image instead of the whole one. Experimental results on the VQA v2.0 validation dataset demonstrate the effectiveness of our proposed attribute network and the constraint between answers and their corresponding attributes, as well as the ability of our method to handle the answers with few training examples.

\end{abstract}

% Note that keywords are not normally used for peerreview papers.
\begin{IEEEkeywords}
Visual Question Answering, Few-shot Learning, Deep Learning
\end{IEEEkeywords}

\IEEEpeerreviewmaketitle

\section{Introduction} \label{introduction}
\IEEEPARstart{T}{he} developments in computer vision and natural language processing enable the machine to deal with complicated tasks that require the integration and understanding of vision and language, e.g. image captioning \cite{anderson2018bottom}, visual grounding \cite{fukui2016multimodal}, visual question answering (VQA) \cite{antol2015vqa, gao2019dynamic, yu2019deep}, and visual dialog \cite{das2017visual, guo2019image}. Compared with image captioning that is to simply describe the topic of an image, VQA needs a complex reasoning process to infer the right answer for a variety of questions. Visual grounding aims to locate the related objects in the image, but VQA takes a further step to convert this information into human language. In addition, VQA is the basic and vital component in visual dialog. Considering the challenges and significance of VQA, increasing research attention has been attracted to it. 

\begin{figure}
	\centering
	\includegraphics[width=1\linewidth]{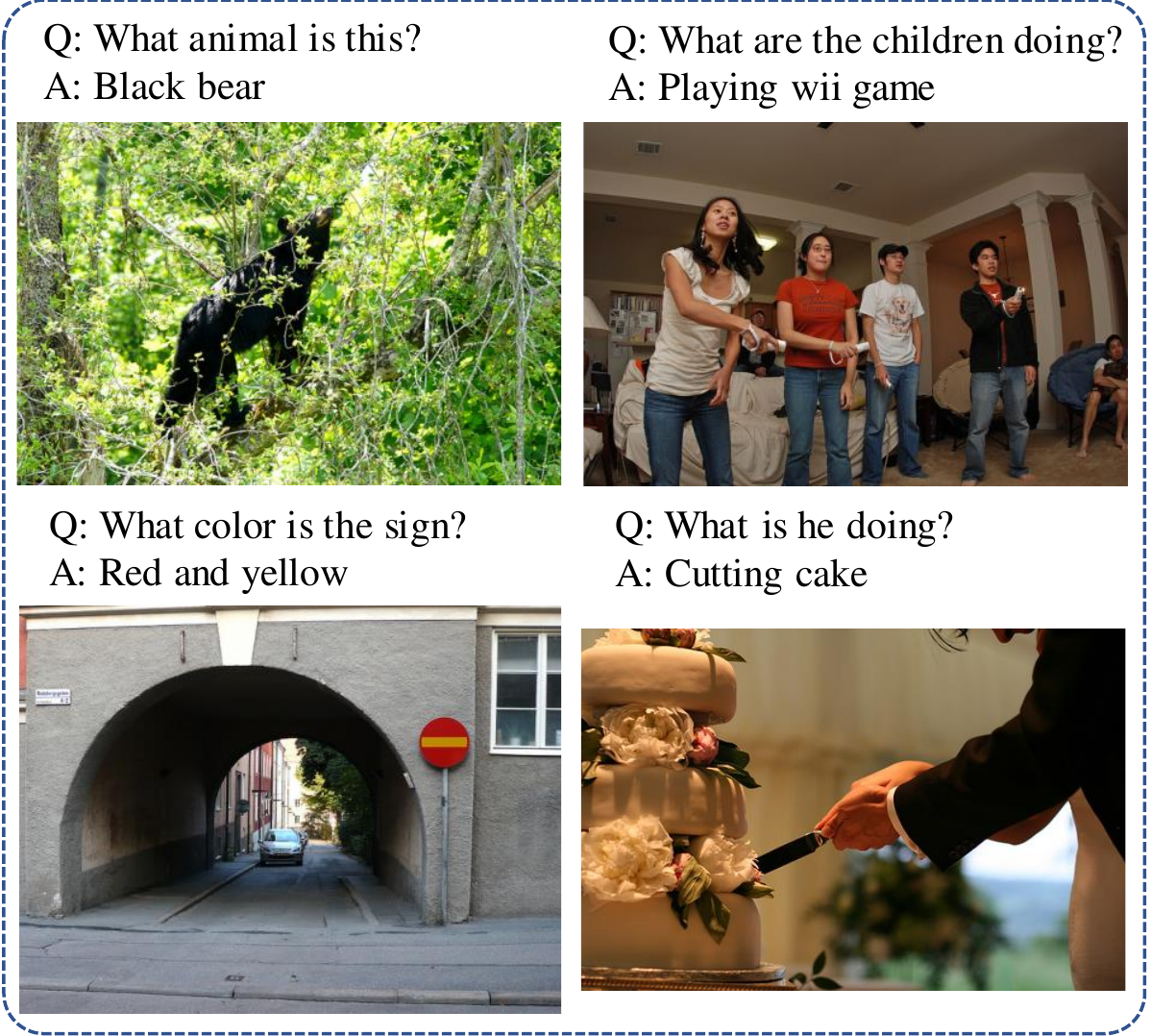}
	\caption{Examples illustrate the compositional answers from VQA v2 \cite{antol2015vqa} dataset.}
	\label{fig:motivation1}
\end{figure}

\begin{figure*}
	\centering
	\includegraphics[width=1\linewidth]{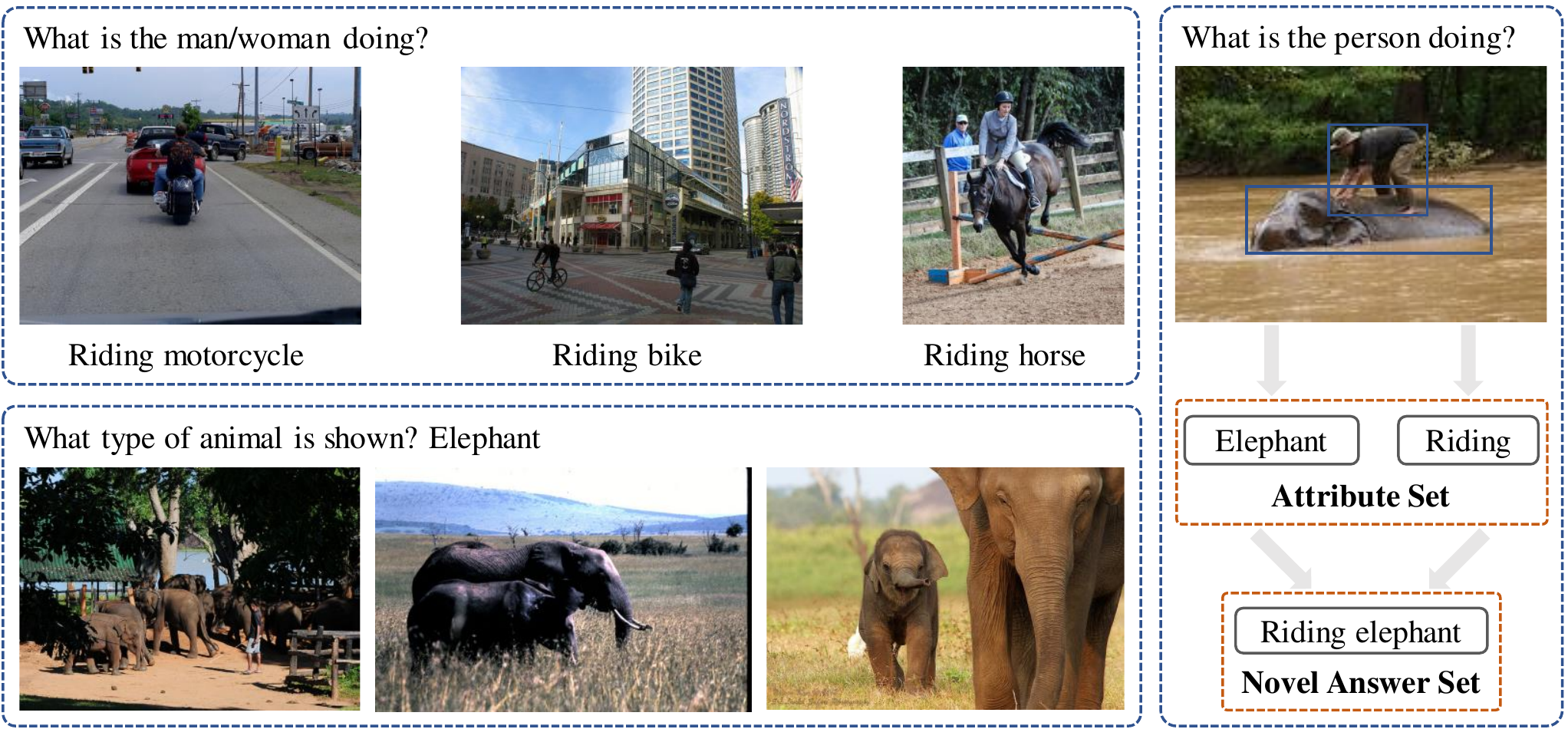}
	\caption{These attributes, such as `riding', `motorcycle', `bike', `horse', and `elephant', found in the base categories can be composed to assist the learning of novel answer `riding elephant'.}
	\label{fig:motivation2}
\end{figure*}

With the understanding of given question by LSTM \cite{hochreiter1997long} or Transformer \cite{vaswani2017attention}, VQA involves many visual tasks from the input image as shown in Figure \ref{fig:motivation1}, such as image recognition \cite{russakovsky2015imagenet} (`black bear' in the top left image), object detection \cite{ren2015faster} (`children' in the top right), scene recognition \cite{zhou2017places}, and human action recognition \cite{gkioxari2018detecting} (`cutting' in the bottom right). However, the categories of these visual tasks are isolated from each other and share few common characters. In contrast, the answers of VQA are compositional and combinations of them, like `cutting cake', `red and yellow', and `playing wii game'. Therefore, it brings VQA a wide range of answers with unbalanced data, and these answers share some common attributes. For example, in Figure \ref{fig:motivation2}, all of these answers in the top left row and right column have the same `riding' action but different objects, among which `riding motorcycle', `riding bike', and `riding horse' are easily recognized by the AI agent due to a number of training examples. Even though there are many examples of `elephant' in the bottom left row, `riding elephant' is still hard to be correctly classified because of few examples. However, humans can learn the novel category only with few glimpses, as we separate this new object into parts that we have ever known. As shown in the right column of Figure \ref{fig:motivation2}, we learn a good view of `riding' with a number of examples in the images of the top row and the meaning of `elephant' in the bottom row, after that, we can easily recognize `riding elephant' by composing the two concepts.

Learning Compositional Representations for Few-shot Recognition \cite{tokmakov2019learning} learns the attributes in the fine-grained dataset \cite{wah2011caltech, xiao2010sun}, but it requires much of the human effort to label the attributes of each image and does not explicitly model the representation of attributes. Visual Question Answering as a Meta Learning Task \cite{teney2018visual} is one of the handful researches for few-shot learning in VQA, it solves this problem from the view of meta-learning, which requires extra memory to store the weights of support set, and it only targets at the `number' questions whose answers only vary from 0 to 6. Following MAML \cite{finn2017model}, Fast Parameter Adaptation \cite{dong2018fast} learns the proper initial weights for the image-text task, but the number of its categories is up to 20-way, limiting its use in reality.

In this paper, we aim to answer the general questions of VQA under the few-shot setting, which are more complex and have a variety of answers. We first create the base set for learning the backbone of our network and the novel set for few-shot answers. Both sets are from the training set of VQA v2.0 and do not need extra human effort to label the attributes of answers. Then we focus on learning the representation of attributes. The visual recognition tasks \cite{russakovsky2015imagenet, tokmakov2019learning} use the mean of vectors (average pooling layer) from Convolutional Neural Network (CNN), such as ResNet \cite{he2016deep}, GoogLeNet \cite{he2015delving}, and ResNext \cite{xie2017aggregated},  for classification, as it describes the whole image, but the attributes come from the different parts of the image, such as `riding' from the upper part and `elephant' from the lower part of the image in Figure \ref{fig:motivation2}. Therefore, we propose the attribute network to make each part of the image score the attributes individually. At last, we model the constraint between the answers and their corresponding attributes in the base set, which later are transferred to learn the representation of few-shot answers in the novel set. 

We conduct experiments on VQA v2.0 dataset. On the validation dataset, which has 1859 novel answers, Our method gains more accuracy than directly classification on the backbone, which is nearly 2\% and 3\% on the 5-shot in top-1 and top-5 respectively. Also, Our model is comparable with \cite{tokmakov2019learning} on the 1-shot and 5-shot, and better in the 10-shot, showing the effectiveness of our network.

\section{Related Work}
In this section, we will first introduce the related research on VQA and then the methods of few-shot learning.

\subsection{Visual Question Answering (VQA)}
VQA is a task to answer the given question based on the input image. The question is usually embedded into a vector with LSTM \cite{hochreiter1997long}, and the image is represented by the fixed-size grid features extracted from a pre-trained model, such as ResNet \cite{he2016deep}. Then both of these features are projected into a joint embedding by addition or concatenation \cite{antol2015vqa}, which is used for answer prediction through a multilayer perceptron (MLP). However, not all features of the image are related to the given question, while some of them should be filtered out before generating the unified vector, therefore attention mechanism is introduced to learn the weight of each grid feature, such as Stack Attention Networks (SAN) \cite{yang2016stacked} and Dual Attention Networks (DAN) \cite{nam2016dual}. 

Due to the different distributions of question and image features, the outer product of both features has a better explanation and performance compared with the linear methods. But because of its high dimension output, it is hard to be optimized. Multimodal Compact Bilinear Pooling (MCB) \cite{fukui2016multimodal} is approaching this process by calculating the count sketch of two features and convolving them in Faster Fourier Transform (FFT) space. Hadamard Product for Low-rank Bilinear Pooling (MLB) \cite{kim2016hadamard} models the common vector with a low-rank matrix by an element-wise multiplication, and Multi-modal Factorized Bilinear Pooling (MFB) \cite{yu2017mfb} increases the rank from 1 to $k$ to accelerate the convergence rate and improve the model's robustness. Furthermore, Bilinear Attention Networks (BAN) \cite{kim2018bilinear} learn the textual and visual attention simultaneously, which builds a mapping from the detected objects of the image to the words of the question. By interpreting BAN as a graph, Bilinear Graph Networks \cite{guo2019bilinear} takes a further step to build the relationships between objects and words by the image-graph as well as between words and words by the question-graph. Motivated by Transformer model \cite{vaswani2017attention}, Dynamic Fusion with Intra- and Inter-modality Attention Flow (DFAF) \cite{gao2019dynamic} and Deep Modular Co-Attention Networks (MCAN) \cite{yu2019deep} involve all the relationships among words and objects, including words and words, words and objects, and objects and objects. Furthermore, Vilbert \cite{lu2019vilbert}, VL-bert \cite{su2019vl}, Structbert \cite{wang2019structbert}, and LXMERT \cite{tan2019lxmert} even directly fine-tune the BERT \cite{devlin2018bert} model without considering the different distributions of question and image features, and this procedure requires a large number of out-domain data, such as image caption dataset \cite{sharma2018conceptual, krishna2017visual, hudson2019gqa} and text-based question answering dataset \cite{rajpurkar2016squad}.

Bottom-up and top-down (BUTD) \cite{anderson2018bottom} focuses on bottom-up attention of image features and proposes a set of salient image regions with natural expression detected by Faster-RCNN \cite{ren2015faster}. Furthermore, its training set contains 1,600 object classes from Visual Genome \cite{krishna2017visual}, larger than the original 80 object classes from COCO \cite{lin2014microsoft}, and it also needs to predict the attributes for the detected objects, such as their action, color, and shape. In Defense of Grid Features (GridFeat) \cite{jiang2020defense} advocates the $1 \times 1$ ROI pooling to speed up extracting the image features while keeping the accuracy.

\subsection{Few-shot Learning}
Few-shot learning targets at recognizing the novel classes with a few labeled examples. The first category of these few-shot algorithms is meta-learning, which learns to learn. Model-Agnostic Meta-Learning (MAML) \cite{finn2017model} learns an easily adaptable model on a variety of tasks, which only needs a few gradient steps to reach a good result. Instead of the Stochastic Gradient Descent optimizer (SGD) \cite{robbins1951stochastic}, Optimization as a Model \cite{ravi2016optimization} proposes a LSTM-based meta-learner for faster training the network. Meta Networks \cite{munkhdalai2017meta} and Visual Question Answering as a Meta Learning Task \cite{teney2018visual} utilize the external memory to update their weight. Another category is metric-learning, which learns to rank the similarity between examples. Siamese neural networks \cite{koch2015siamese} classify the unseen image by comparing it with the labeled ones. Matching Networks \cite{vinyals2016matching} replace the softmax with the cosine similarity in computing the distance between the query feature and support features, while Prototypical Networks \cite{snell2017prototypical} use the Euclidean distance between the query feature and the mean of support features. Few-shot Learning with Graph Neural Networks \cite{garcia2017few} build a graph on the labeled and unlabeled examples. Apart from these two categories, data augmentation methods, such as \cite{hariharan2017low, antoniou2017data} learn a generator in the base data and use it to hallucinate the new novel data. Moreover, A Closer Look at Few-shot Classification \cite{chen2019closer} compares these representative algorithms and develops a simple baseline approach on top of the frozen CNN, which achieves state-of-the-art on both mini-ImageNet \cite{vinyals2016matching} and CUB \cite{wah2011caltech} datasets. 

\subsection{Learning with attributes}
Learning the compositional attributes is a natural way for humans to learn the few-shot and zero-shot examples. With modeling the transformation network of these compositional attributes, From Red Wine to Red Tomato \cite{misra2017red} can category the unseen combination of concepts. Measuring compositionality in representation learning \cite{andreas2019measuring} builds a new evaluation metric to estimate the sum of attributes that describes the input image. Dense Attribute-Based Attention\cite{huynh2020fine} assigns the region features from the image to their semantic vectors from GloVe \cite{pennington2014glove}. Learning Compositional Representations \cite{tokmakov2019learning} disentangles the feature space of categories into sub-spaces corresponding to their attributes, which is similar to our method, but it omits that each part of the encoded features makes a different contribution to the attributes. 

\section{Methods}
The goal of VQA task is to answer the given question $T$ based on the input image $I$. With the object-detector Faster-RCNN \cite{ren2015faster, anderson2018bottom, jiang2020defense}, we convert the input image $I$ into object features $V = (v_1, \dots, v_n)$ with $v_i \in R^D$, where $n$ is the number of detected objects, and $D$ is the feature dimension. The question $(t_1, \dots, t_m)$ is a sequence of $m$ words. It can be encoded using LSTM \cite{hochreiter1997long} to $Q = (q_1, \dots, q_m)$, where $Q = \text{LSTM}(T)$, $Q \in R^{C \times m}$, and $C$ is the dimension of output features. We can get the joint embedding of $Q$ and $V$ with these well-known VQA models, such as MLB \cite{kim2016hadamard}, MFB \cite{yu2017mfb}, BAN \cite{kim2018bilinear}, and MCAN \cite{yu2019deep}. However, BGN \cite{guo2019bilinear} has a better explanation and performance than them, and it can be easily trained with less time and less GPU resource than these BERT-base models needed to be fine-tuned, such as Vilbert \cite{lu2019vilbert}, VL-bert \cite{su2019vl}, and LXMERT \cite{tan2019lxmert}. Thus it is chosen as the backbone of our framework. 

\begin{figure*}
	\centering
	\includegraphics[width=1\linewidth]{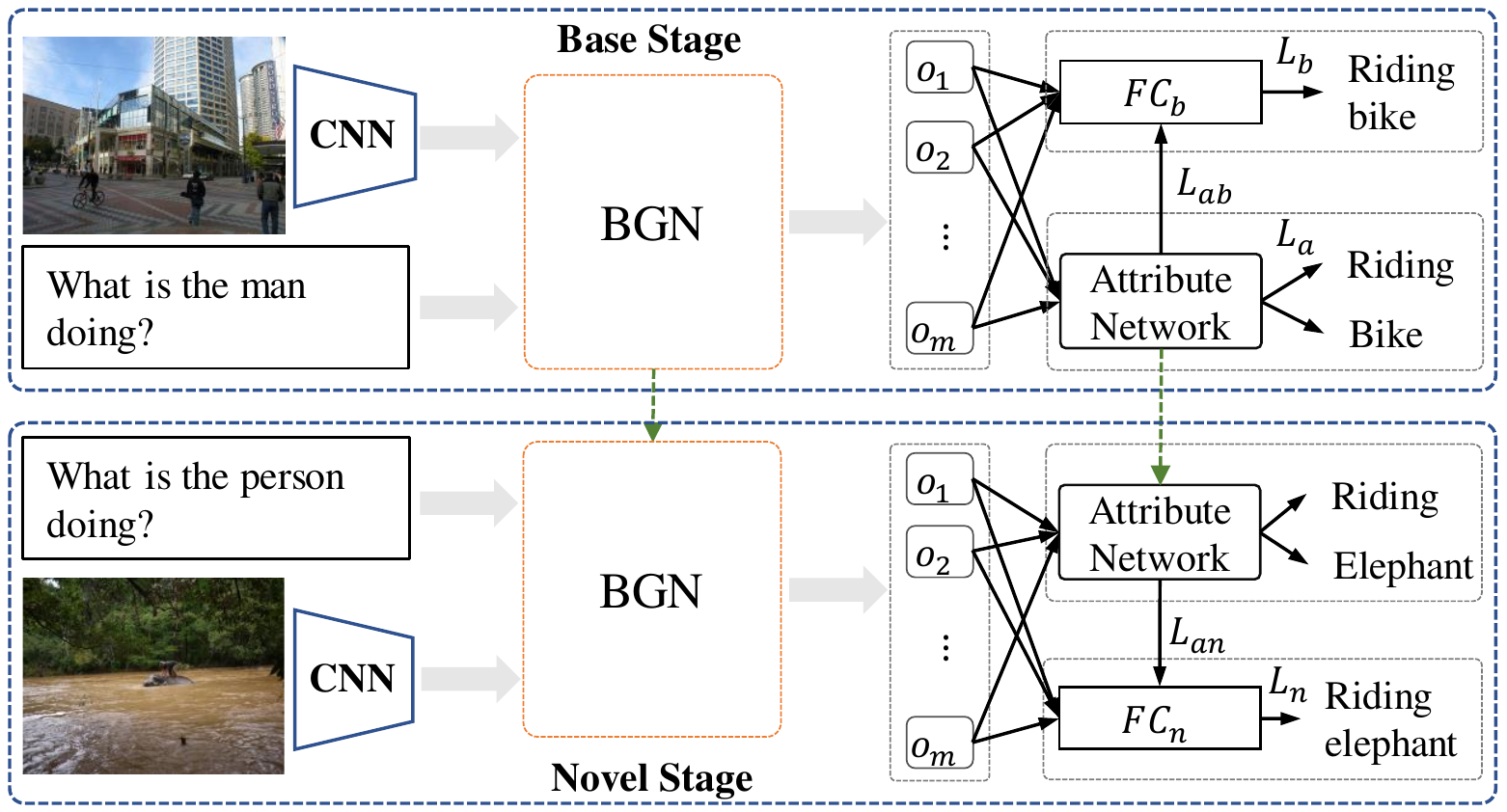}
	\caption{The framework of our model. In the base stage, we first model the relationship between words and objects with BGN \cite{guo2019bilinear}, then we learn the vectors of base answers $A_b$ and attributes $U$ with the attribute network. In the novel stage, we utilize the well-trained embeddings of attributes to improve the representation of novel answers $A_n$.}   
	\label{fig:framework}
\end{figure*}

\subsection{Bilinear Graph Network}
BGN models the relationship between words and objects with the image-graph (bias terms are omitted without loss of generality):
\begin{equation}
G_{j}^{e} =\mathrm{softmax}\Big((((\mathds{1} \cdot \mathbf{p}_{j}^{e\top}) \circ \sigma(Q^\top\mathbf{U'}^{e})) \sigma(V^\top \mathbf{V'}^e)^\top \Big), 
\end{equation}
where $G_e \in R^{m \times n \times g^e}$ is the graph attention weight between $Q$ and $V$, $g^e$ is the number of glimpses, and $\sigma$ is the ReLU activation function denoted as $\sigma(x)=\max(x,0)$. The parameters $\mathbf{U'}^e \in R^{C \times K}$ and $\mathbf{V'}^e \in R^{D \times K}$ are shared among glimpses except for $\mathbf{p}_j^e \in R^K$. After learning the graph attention, it generates the joint embeddings as:
\begin{equation}
H_j' = W^e_j(\sigma(H_{j-1}^\top \mathbf{U}_j^e) \circ G^e_j \sigma(V^\top \mathbf{V}_j^e))^\top + H_{j-1}',
\end{equation}
where $\mathbf{U}_j^e \in R^{C \times K}$, $\mathbf{V}_j^e \in R^{D \times K}$, $W^e_j \in R^{C \times K}$ projects the joint embeddings to the same dimension of $Q$ for residual connection following \cite{kim2018bilinear}. $H_j' \in R^{K \times m}$ represents the output of image-graph at glimpse $j$ with $H_0 = Q$, and we use the output of last glimpse to represent the whole image-graph, denoted as $H = H_{g^e}'$.

BGN also builds the graph between these joint embeddings in term of words to exchange context information:
\begin{align}
G_{j}^{r} &=\mathrm{softmax}\Big((((\mathds{1} \cdot \mathbf{p}_{j}^{r\top}) \circ \sigma(H^\top\mathbf{U'}^{r})) \sigma(H^\top \mathbf{V'}^r)^\top \Big), \\
O_j' &= W^r_j(\sigma(O_{j-1}^\top \mathbf{U}_j^r) \circ G^r_j \sigma(H^\top \mathbf{V}_j^r))^\top + O_{j-1}',
\end{align}
where $G_r \in R^{m \times m \times g^r}, \mathbf{U'}^r, \mathbf{V'}^r, \mathbf{U}_j^r,  \mathbf{V}_j^r, W^r_j \in R^{C \times K}$, $\mathbf{p}_j^r \in R^K$, and $O_0' = H$. The outputs of question-graph $O$, abbreviated version of $O_{g^r}'$, can be utilized to answer the question by summarizing all the nodes to represent the whole graph:
\begin{equation} \label{eq:g6}
z = \sum_{i=1}^{m} o_i.
\end{equation}

\subsection{Learning Compositional Representation}
Based on the number of training examples of answers $A$, we classify them into two sets, the base set $A_b$ with a large number of training examples and the novel one $A_n$ with only a few examples, denoting that $A = A_b \cup A_n$ and $A_b \cap A_n = \emptyset$. 

The naive way of computing the score between answers and joint embedding of words and objects is formulated as:
\begin{equation}
S = W^{b'\top}\sigma(W^b z),
\end{equation}
where $W^b \in R^{2C \times C}, W^{b'} \in R^{2C \times |A_b|}$, and $|A_b|$ is the size of base answer set $A_b$. Moreover, $w^{b'}_i \in W^{b'}$ represents the vector of answer $a_i \in A_b$. Furthermore, there might exist multiple correct answers for a pair of question and image, so we utilize the binary cross-entropy loss (BCE) as loss function, which is calculated as:
\begin{equation} \label{eq:base_loss}
L_b = -\sum_{i = 1}^{|A_b|}(y_i \log \phi(s_i) + (1-y_i)\log(1 - \phi(s_i))),
\end{equation}
where $s_i \in S$ is the score of answer $a_i$, $\phi(x)$ is the sigmoid function denoted as $\phi(x)=\frac{1}{1 + e^{-x}}$, and $y_i= \min(\frac{\text{number of people that provided answer} ~a_i}{3}, 1)$. 

As we mentioned in Section \ref{introduction}, answer $a_i \in A$ is composed of several attributes, denoting as $a^u_i = \{u_{i,1}, \dots, u_{i,k_i}\}$, where $u_{i,j} \in U$, $U$ is the attributes set, and $k_i$ is the number of attributes in answer $a_i$. 

\begin{figure}
	\centering
	\includegraphics[width=0.67\linewidth]{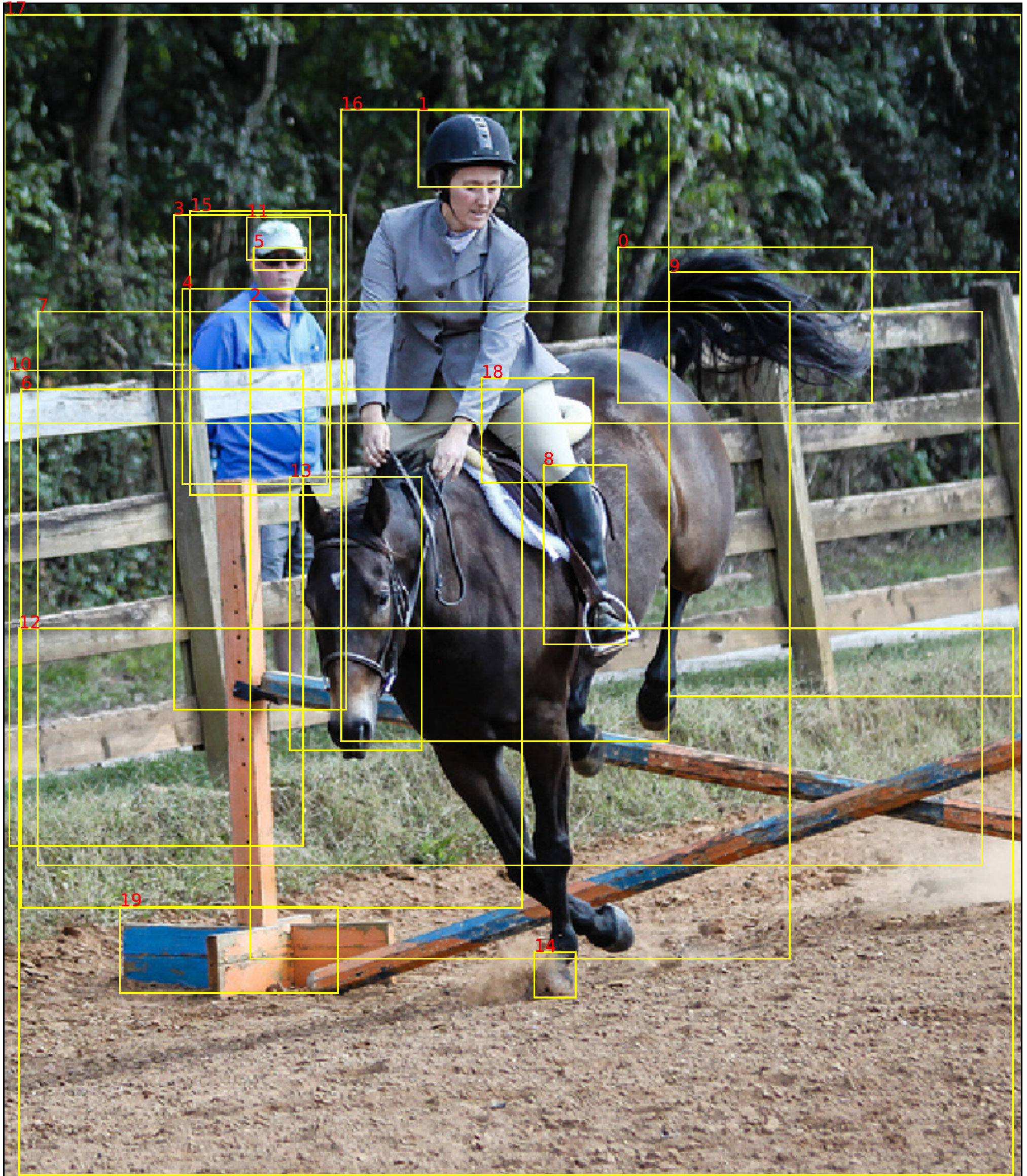}
	\includegraphics[width=0.29\linewidth]{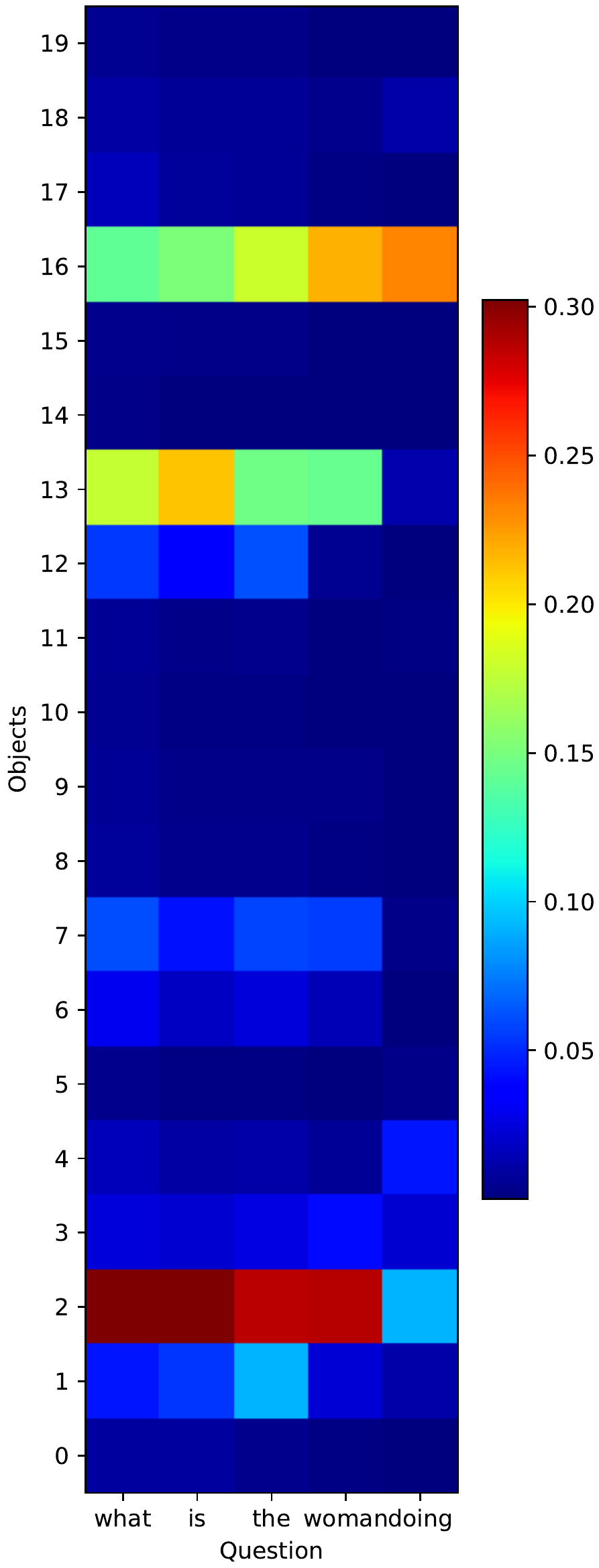}
	\caption{Attention map between the words of question and the objects of image.}
	\label{fig:attention_map}
\end{figure}
\textbf{Attribute Network}: The key of learning compositional answers is to generate a good representation of their attributes, and we can also use the unified vector of graph like $z$ to learn these attributes. But the attributes in the right answer may come from the different parts of the given image and should be handled separately. For example, in Figure \ref{fig:attention_map}, the phrase `what is' focuses on the object of `horse' (object 2), while `woman doing' pays more attention to the object of `woman' (object 16). Therefore, we propose the attribute network to compute the score between each node of the graph and the attributes respectively instead of a whole one:
\begin{equation}
S^a = \sum_{i=1}^{m} W^{a'\top}\sigma(W^a o_i),
\end{equation}
where $W^a \in R^{2C \times C}, W^{a'} \in R^{2C \times |U|}$, and $w^{a'}_i \in W^{a'}$ represents the embedding of attribute $u_i \in U$. The non-linear operator $\sigma$ used before summarization of graph nodes makes the representation of attributes more independent on each other than after it.

Due to that several attributes comprise the answer, so we also utilize BCE as its loss function, which is calculated as:
\begin{equation}
L_a = -\sum_{i = 1}^{|U|}(y^a_i \log \phi(s^a_i) + (1-y^a_i)\log(1 - \phi(s^a_i))),
\end{equation}
where $y^a_i = max(1,\sum_{u_i \in a^u_j} y_j)$ is the label for attribute $u_i$. 

Moreover, the answer can be reconstructed from the representation of its attributes, so we minimize the distance between them:
\begin{equation}
L_{ab} = \sum_{i = 1}^{|A_b|} f(w^{b'}_i, \sum_{u_j \in a^u_i}w^{a'}_j),
\end{equation}
where $f$ is the distance function, such as cosine similarity and mean square error (MSE). Overall, we have all the loss during training in the base stage to learn the vectors of answers and attributes, as well as the backbone of our framework as shown in the upper part of Figure \ref{fig:framework}:
\begin{equation}
L_{base} = L_a + L_b + \lambda L_{ab},
\end{equation}
where $\lambda$ is the hyper-parameter that adjusts the importance of distance loss.

\textbf{Constrained by Attributes}: The representation of answers is hard to be learned in the novel dataset, since there are only a few examples for them, resulting in easily over-fitting. Nevertheless, the attributes of them have been obtained in the base stage with our attribute network and can assist in learning the embeddings of these answers as shown in the lower part of Figure \ref{fig:framework}:
\begin{align}
&L_n  = -log\frac{\exp(w^{n'\top}_i\sigma(W^b z))}{\sum_{j=1}^{|A_n|} \exp(w^{n'\top}_j\sigma(W^b z))}, \\
&L_{an} = \sum_{i = 1}^{|A_n|} f(w^{n'}_i, \sum_{u_j \in a^u_i}w^{a'}_j), \\
&L_{novel} = L_n + \lambda L_{an},
\end{align}
where $w^{n'}_j \in R^{2c}$ is the vector of novel answer $a_j \in A_n$, and we sample only one answer $a_i$ from the right ones based on the setting of few-shot learning. Note that, we only compute the gradient of $w^{n'}$ in the loss of $L_{an}$ and $L_n$ for updating its weights, while keeping the parameters of attributes fixed to regularize the learning of novel answers.

\section{Experiments}
In this section, we evaluate our method on VQA v2.0 dataset \cite{goyal2017making}. We first introduce this dataset following by constructing the few-shot dataset and then describe our implementation details and results, and finally the qualitative analysis.

\subsection{VQA v2.0 Datasets}
The dataset was built based on the MSCOCO images \cite{lin2014microsoft}, and it contains 1.1M questions asked by human and each question is annotated by 10 people. Compared with VQA v1.0 dataset \cite{antol2015vqa}, it emphasizes the visual understanding by reducing the text bias learned from the questions. Compared with GQA \cite{hudson2019gqa} and Visual Genome \cite{krishna2017visual}, its answers are compositional and more complex, which leads to a large number of candidates and causes the problem of few training examples. The dataset is split into three parts: training, validation, and test, and the answers of the training and validation dataset are published for training model, while those of the test dataset are unknown and should be predicted by the proposed model before being uploaded to the server for performance evaluation. 

\textbf{Constructing the few-shot dataset}: The answers of VQA dataset can be classified into three types, i.e. `yes/no', `number', and `others', and we focus on the ones of `others', as they are more complex and diverse than the other two categories. We first collect our base answer set with these answers appearing in the training dataset more than 40 times, resulting in $|A_b| = 2584$. Different from the strategy used in \cite{teney2018tips, yu2019deep}, which collect only one correct answer for each question, we gather multiple ones that are ever responded. Then we use spacy \footnote{https://spacy.io/models/en} to separate these answers into tokens, which are regarded as the attributes of them and organized into the attribute set after removing some meaningless words (`and', `is', 'do', '\&', and `are'), leading to $|U| = 2108$ attributes. By composing these attributes, we have the novel answer set $A_n$ with answers that are not in $A_b$ and appear more than 10 times, causing 1859 answers, and similar procedures are also applied on the validation dataset for evaluation. Table \ref{table:dataset} demonstrates the number of questions and answers for both datasets.

\begin{table}
	\centering
	\caption{Number of questions and answers for the split answer set.}
	\label{table:dataset}
	\resizebox{0.9\linewidth}{!}{
		\begin{tabular}{lcccc}
			\hline
			\textbf{Split} & \textbf{Train} & \textbf{Val} & \textbf{Answer} & \textbf {times}\\
			\hline
			Base & 219269 & 105679 & 2584 & $t \ge 40$ \\
			Novel & 27382 & 12855 & 1859 & $10\le t <40$ \\
			\hline
	\end{tabular}}
\end{table}

\subsection{Implementation Details}
Following BGN \cite{guo2019bilinear}, we truncate or pad the length of question $m$ to 15 words, and the input dimension of LSTM is 600, 300 of which is learned by our model and another 300 is from pre-trained GloVe vector \cite{pennington2014glove}, and the output dimension $C$ is $1,024$. Note that BERT \cite{devlin2018bert} can also be used as a question encoder to improve the performance of BGN. The object features are extracted from a Faster-RCNN model pre-trained on Visual Genome \cite{anderson2018bottom}, getting $n = 100$ objects for each image and $D = 2,048$ for each object. Though stacking more layers of BGN can achieve a better result, it is not our contribution and consumes more time in training and evaluating the model, thus we choose the one-layer BGN as our backbone with glimpses of $g^r=g^e=4$ for both the image-graph and question-graph. The size of joint embedding $K$ is set to $1,024$. Weight Normalization \cite{salimans2016weight} and Dropout \cite{srivastava2014dropout} with $p = 0.2$ are added after each linear mapping to stable the output and prevent from over-fitting.

We implement our model on Pytorch \footnote{https://pytorch.org/} v1.2, the batch size is set to 128, and Adamax \cite{kingma2014adam} is used to optimize our model. In the base stage, we train the BGN model, the representation of base answers $\text{FC}_b$, and attribute network from scratch with data whose answers are in the base set. The initial learning rate is 0.001 and grows by 0.001 every epoch until reaching 0.004 for a warm start, keeps constant until the eleventh epoch, and decays by 1/4 every two epochs to 0.00025. In the novel stage, we only train the vectors of novel answers on top of the frozen BGN and attribute network, its learning rate is set to 0.004 for fast convergence until 55, 35, and 20 epochs for 1-, 5-, and 10-shot learning respectively, then following by 5 epochs with 0.001. 

BGN is also chosen as our baseline model. In the novel stage, we just replace the FC layer of base answers with the one of novel answers and learn the representation of them with the fixed backbone. Due to easily over-fitting on the novel few-shot dataset, we reduce the training epochs of it to 40, 20, and 14 for 1-, 5-, and 10-shot learning respectively.

\subsection{Evaluation Metrics}
We aim to solve the general VQA problems as other similar models \cite{yu2019deep, guo2019bilinear, anderson2018bottom, yu2017mfb, lu2019vilbert}, thus we evaluate the top-1 accuracy on the validation dataset of VQA v2 under the 1-, 5-, and 10-shot settings by tools from \cite{antol2015vqa} \footnote{https://github.com/GT-Vision-Lab/VQA}. Additionally, we considerate the top-5 performance following \cite{tokmakov2019learning}, which is the best score among the top-5 predicted answers.

\subsection{Ablation Study}
\begin{table}
	\centering
	\caption{Influence of $\lambda$ and initialization methods on our models.}
	\label{table:lambda}
	\resizebox{1.0\linewidth}{!}{
		\begin{tabular}{l|c|c|c|c|c|c|c|}
			&  & \multicolumn{3}{c|}{Top-1} & \multicolumn{3}{c|}{Top-5} \\
			\textbf{$\lambda$} & Init & \textbf{1-shot} & \textbf{5-shot} & \textbf{10-shot} & \textbf{1-shot} & \textbf{5-shot} & \textbf{10-shot} \\
			\hline
			0.0 & He & 6.45 & 12.53 & 14.31 & 12.59 & 23.36 & 25.88 \\
			0.1 & He & 7.38 & 14.37 & \textbf{15.46} & 14.84 & \textbf{26.24} & \textbf{27.37} \\
			0.2 & He & 8.16 & \textbf{14.53} & 15.19 & 17.47 & 26.11 & 26.76 \\
			0.5 & He & 9.18 & 13.53 & 13.70 & \textbf{19.76} & 24.47 & 24.60 \\	
			0.1 & Attribute & \textbf{11.13} & 13.64 & 14.64 & 22.23 & 25.02 & 25.94 \\
	\end{tabular}}
\end{table}

We conduct the ablation studies to verify the contribution of our attribute network in Table \ref{table:lambda}. Note that when $\lambda$ becomes zero, our network is the same as BGN. The performance grows with the increase of $\lambda$ on the 1-shot learning for both the top-1 and top-5 accuracy, showing the effectiveness of our proposed network. Nevertheless, the performance drops on the 5-shot and 10-shot learning when $\lambda > 0.2$, it can be explained that the answers can be represented by the summing of embeddings of attributes to some extent with $\lambda = 0.1$ and $\lambda = 0.2$, but a higher value like $\lambda = 0.5$ limits its potential to learn new information from the domain of few-shot dataset. This phenomenon is also found in the last line, which directly initializes the embedding of answer from its attributes instead of He initialization \cite{he2015delving}, the accuracy improves a lot on the 1-shot learning but drops rapidly on the two other settings. Therefore, we choose $\lambda = 0.1$ as our parameter and He initialization as the initialization method. 

\subsection{Comparison with Other Models}
\begin{table}
	\centering
	\caption{Top-1 and top-5 Accuracy on the novel answer set of VQA v2 validation dataset.}
	\label{table:overall}
	\resizebox{1.0\linewidth}{!}{
		\begin{tabular}{l|c|c|c|c|c|c|}
			& \multicolumn{3}{c|}{Top-1} & \multicolumn{3}{c|}{Top-5} \\
			\textbf{Model} & \textbf{1-shot} & \textbf{5-shot} & \textbf{10-shot} & \textbf{1-shot} & \textbf{5-shot} & \textbf{10-shot} \\
			\hline
			BGN \cite{guo2019bilinear} & 6.45 & 12.53 & 14.31 & 12.59 & 23.36 & 25.88 \\
			BGN + SUM & 6.27 & 14.06 & 15.15 & 13.33 & 26.19 & 26.59 \\
			BGN + LCR \cite{tokmakov2019learning} & 7.20 & 14.39 & 15.19 & 15.92 & 26.12 & 26.85 \\
			BGN + AN (ours) & 7.38 & 14.37 & 15.46 & 14.84 & 26.24 & 27.37 \\
	\end{tabular}}
\end{table}

In Table \ref{table:overall}, we compare our method with others: \textbf{BGN} trains the weights of novel answers from scratch; \textbf{BGN + SUM} directly utilizes the joint embedding $z$ to learn the weights of novel answers with the constraint between answers and their attributes; \textbf{BGN + LCR} is similar to ours, except that it uses the cosine similarity for the distance function $f$ and does not involve the network to model the loss between attributes and representation of inputs (image for \cite{tokmakov2019learning} and image and question for VQA). 

The last three lines of Table \ref{table:overall} show that the extra module on modeling the constraint between the categories and their corresponding attributes can improve the performance of few-shot learning compared with BGN. Moreover, the performance of our BGN + AN model is comparable to BGN + LCR \cite{tokmakov2019learning} in the 1- and 5-shot learning, and much better than it in the 10-shot one. Since LCR does not consider the relationship between the joint embedding and the attributes, such as $L_A$ in Figure \ref{fig:framework}. Given more training data, our model learns a better representation of the attributes in the base stage, which later is used to improve the training of categories in the novel stage. Furthermore, with the same backbone, the proposed attribute network in BGN + AN works better than the sum method in BGN + SUM. As these attributes are from the different parts of the given image, such as `elephant' from the lower part and `riding' from the upper one shown in Figure \ref{fig:motivation2}. Accordingly, each of the output nodes $O$ of the question-graph has both textual information from the question and related visual information from the image, thus it can be used to represent the attribute in the attended part of the image, proving that simply summarizing of these nodes like other general recognition task \cite{tokmakov2019learning, he2016deep} are not proper to represent all the attributes.

\subsection{Qualitative Analysis}
\begin{figure*}
	\centering
	\includegraphics[width=1\linewidth]{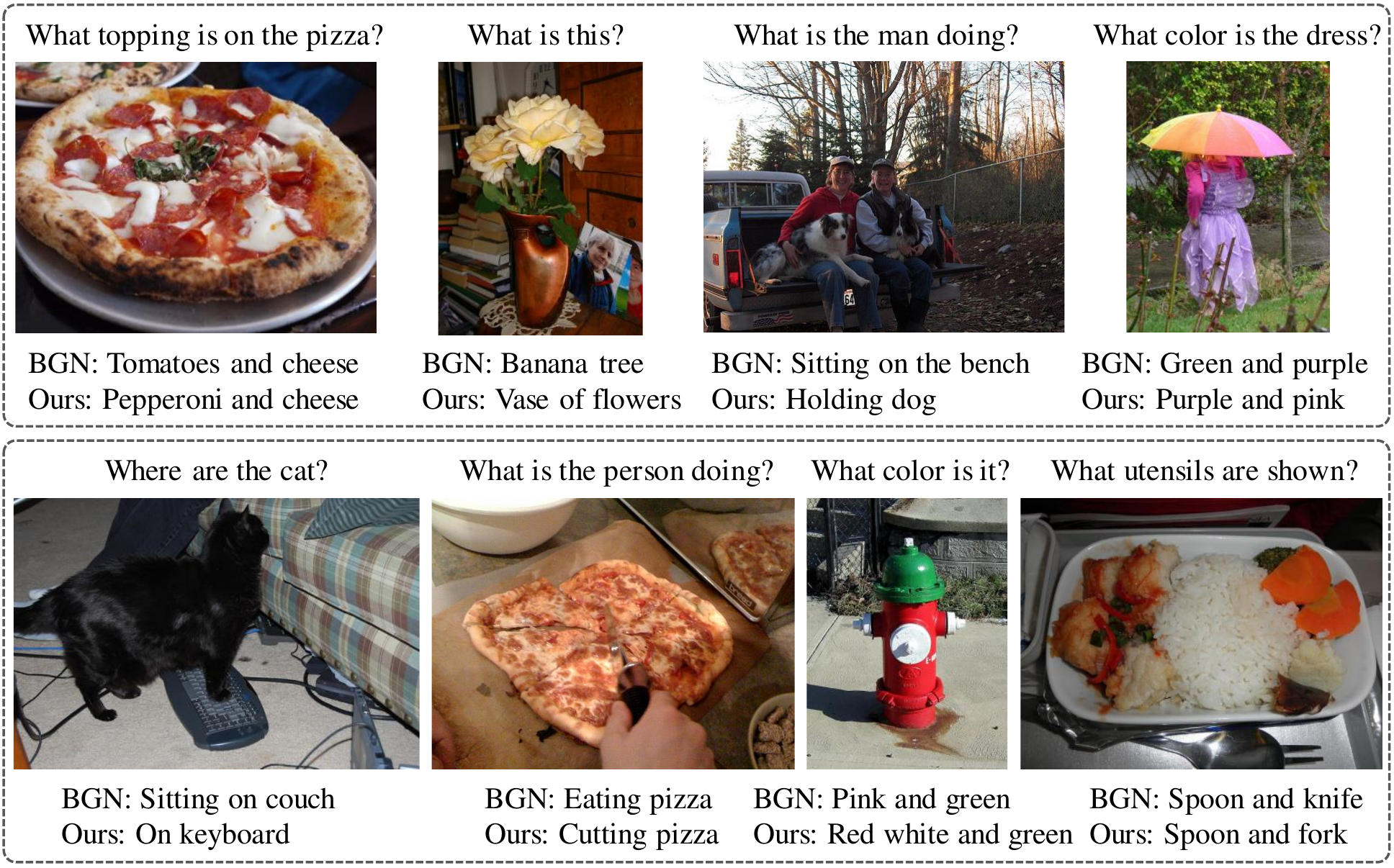}
	\caption{Examples illustrate the compositional answers from the validation dataset of VQA v2.}
	\label{fig:visual}
\end{figure*}

\begin{table}
	\centering
	\caption{Categories of questions whose answers are `pizza' in the base dataset.}
	\label{table:pizza}
	\resizebox{0.7\linewidth}{!}{
	\begin{tabular}{c|c|c|c|}
	\textbf{Total} & \textbf{Eating} & \textbf{Cutting} & \textbf{Other} \\
	\hline
	1633 & 299 & 30 & 1304
	\end{tabular}}
\end{table}

To visualize the effects of our method, we present the answers predicted by BGN and our BGN + AN network in Figure \ref{fig:visual}. In the second image of the bottom row, BGN cannot correctly answer the question. Because in the base stage, when it comes up with `pizza' in the image that appears in the right answer, it is always related to `eating' in the question, which is a strong bias learned by the BGN model as shown in Table \ref{table:pizza}. In contrast, our model learns the `cutting' action from other examples besides `pizza', like `cake', `apple', and `paper'. In the novel stage, the attributes of `cutting' and `pizza' detected in the given image are used to improve the representation of the answer `cutting pizza'. Similar problems are also found in the third image of the top row (people sit on the bench) and the first image of the bottom row (cat sits on the couch). Moreover, an object is always pictured with multiple colors, such as the fourth image of the top row and the third image of the bottom row. Apart from learning the vector of new combination color from scratch in the novel stage, we utilize the well-trained representation of the single color in the base stage to enhance it without extra human effort on annotation. This is also applied to find the objects in the image, such as vase and flowers in the second image of the top row and spoon and fork in the fourth image of the bottom row.

\section{Conclusions}
In this work, we apply the few-shot setting to the general VQA problem, then we propose a two-stage network to overcome it. We advocate the few-shot dataset of VQA, which has two subsets, namely the base set and novel set, and we generate the attributes of answers without human effort. Given a large amount of training data from the base dataset, we utilize BGN to model the relationship between words from the question and objects from the image and generate their joint embeddings in the base stage. Also, we get the representation of attributes with the designed attribute network and with the constraint between attributes and their source answers. With the assistant of well-trained vectors of attributes, we improve the representation of compositional answers that have only a few examples in the novel stage. Applying our approach to the validation set of VQA v2.0, our method has a better performance than the baseline method and other similar networks in the top-1 and top-5 accuracy. However, the result (15.46\% on 10-shot in top-1 accuracy) is far from satisfactory compared with it on the base set (61.18\% on BGN), and more research should be further investigated on this problem.

\bibliographystyle{IEEEtran}
\bibliography{main}

\begin{IEEEbiography}[{\includegraphics[width=1in,height=1.25in,clip,keepaspectratio]{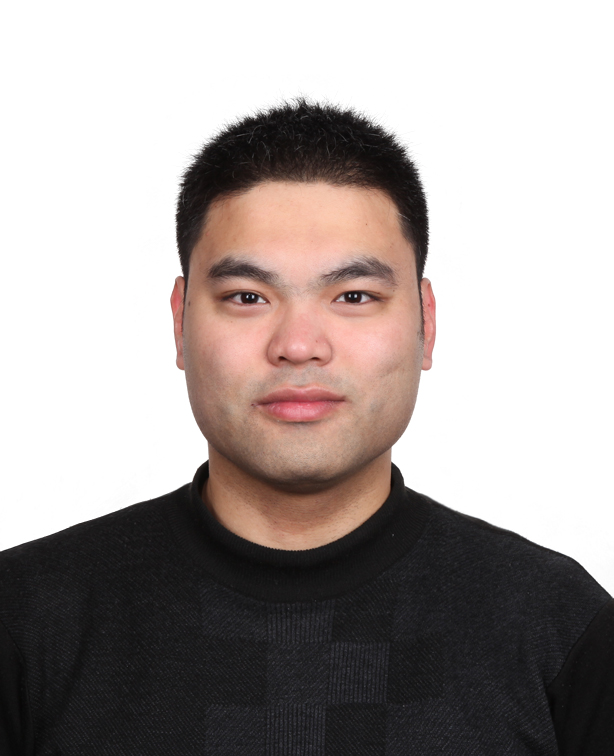}}]{Dalu Guo} obtained his Bachelor and Master degree in the School of Compute Science and Engineering from Beihang University, Beijing, China. He is currently a Ph.D student at UBTECH Sydney Artificial Intelligence Centre and the School of Computer Science, the Faculty of Engineering and Information Technologies, the University of Sydney. His research interests mainly include visual question answering, visual dialog, natural language processing, and so on. His papers has been published in IEEE CVPR, and he got the first place in the Visual Dialog Challenge 2018 (SiVL Workshop of ECCV 2018) and second place in the VQA Challenge 2020 (Visual Question Answering and Dialog Workshop of CVPR 2020).
\end{IEEEbiography}

\begin{IEEEbiography}[{\includegraphics[width=1in,height=1.25in,clip,keepaspectratio]{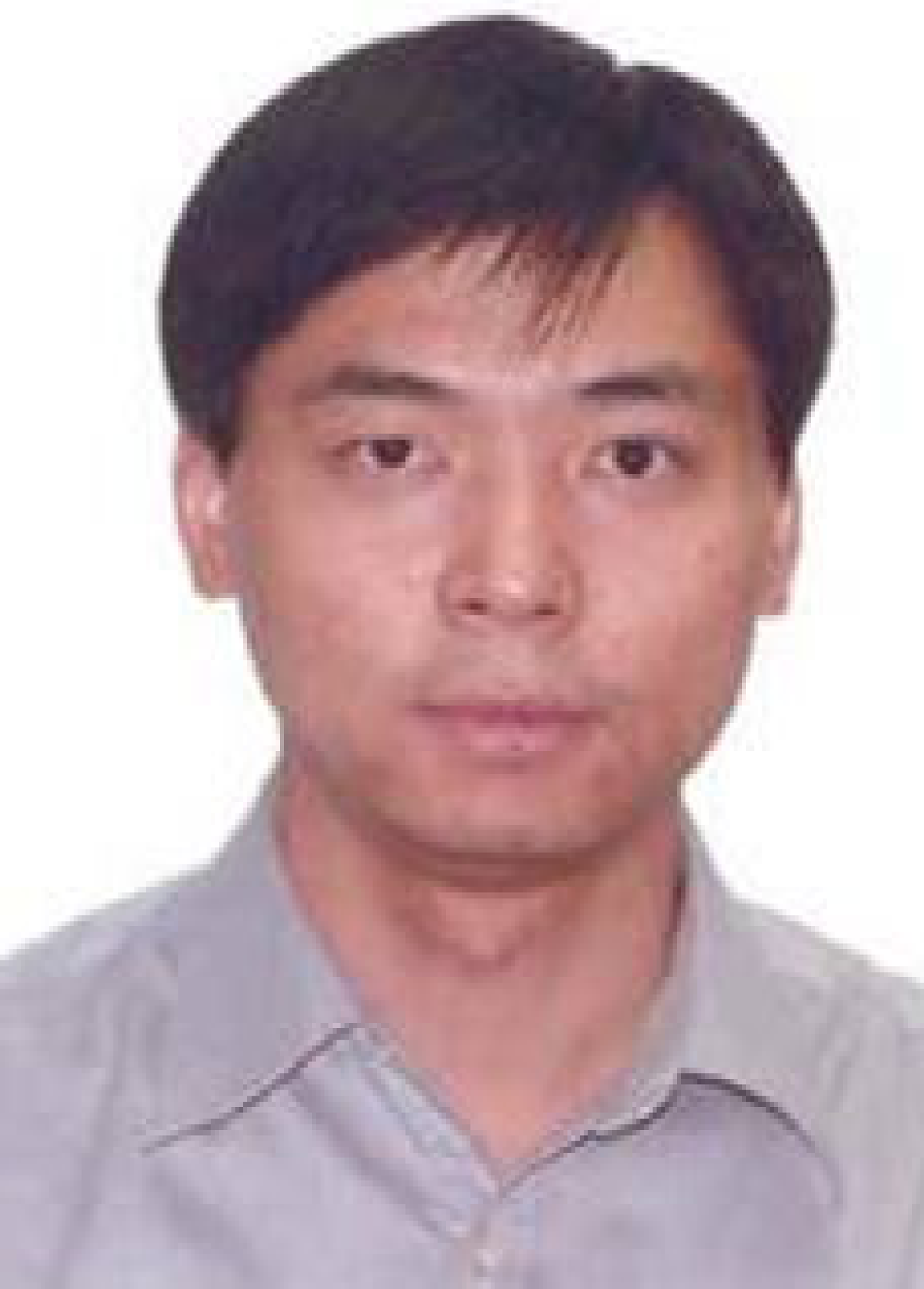}}]{Dacheng Tao} (F’15) is Professor of Computer Science and ARC Laureate Fellow in the School of Computer Science and the Faculty of Engineering, and the Inaugural Director of the UBTECH Sydney Artificial Intelligence Centre, at The University of Sydney. His research results in artificial intelligence have expounded in one monograph and 200+ publications at prestigious journals and prominent conferences, such as IEEE T-PAMI, T-NNLS, IJCV, JMLR, AAAI, IJCAI, NIPS, ICML, CVPR, ICCV, ECCV, ICDM, and KDD, with several best paper awards. He received the 2018 IEEE ICDM Research Contributions Award. He is a Fellow of the IEEE and the Australian Academy of Science.
\end{IEEEbiography}

\end{document}